# Airway Skill Assessment with Spatiotemporal Attention Mechanisms Using Human Gaze

**Jean-Paul Ainam, Rahul**
**CeMSIM, RPI**
**Troy, New York**
**ainamj@rpi.edu, rahul@rpi.edu**

**Lora Cavuoto**
**University at Buffalo**
**Buffalo, NY**
**loracavu@buffalo.edu**

**Matthew Hackett, Jack Norfleet**
**CCDC, Center STTC**
**Orlando, FL**
**matthew.g.hackett.civ@army.mil,**
**jack.e.norfleet.civ@army.mil**

**Suvranu De**
**Florida State University**
**Tallahassee, FL**
**sde@eng.famu.fsu.edu**

## ABSTRACT

Airway management skills are critical in emergency medicine and are typically assessed through subjective evaluation, often failing to gauge competency in real-world scenarios. This paper proposes a machine learning-based approach for assessing airway skills, specifically endotracheal intubation (ETI), using human gaze data and video recordings. The proposed system leverages an attention mechanism guided by the human gaze to enhance the recognition of successful and unsuccessful ETI procedures. Visual masks were created from gaze points to guide the model in focusing on task-relevant areas, reducing irrelevant features. An autoencoder network extracts features from the videos, while an attention module generates attention from the visual masks, and a classifier outputs a classification score. This method, the first to use human gaze for ETI, demonstrates improved accuracy and efficiency over traditional methods. The integration of human gaze data not only enhances model performance but also offers a robust, objective assessment tool for clinical skills, particularly in high-stress environments such as military settings. The results show improvements in prediction accuracy, sensitivity, and trustworthiness, highlighting the potential for this approach to improve clinical training and patient outcomes in emergency medicine.

## ABOUT THE AUTHORS

**Jean-Paul Ainam, PhD** is a research scientist at the Center for Modeling, Simulation & Imaging in Medicine, RPI, focusing on Machine learning applied to videos. His research interests include Generative Adversarial Networks, Scene understanding, and deep neural network architecture design for multi-view and multimodal data.

**Rahul** is an Assistant Professor in the Department of Biomedical Engineering at Rensselaer Polytechnic (RPI). He is actively engaged in interdisciplinary research that combines physics, biomedical imaging, and data-driven analytics for solving problems in healthcare with the eventual goal of reducing medical errors and advancing patient safety.

**Lora Cavuoto, PhD** is a Professor in the Department of Industrial and Systems Engineering at the University at Buffalo. Dr. Cavuoto's research focuses on improving occupational safety and productivity through ergonomic principles and advanced technologies.

**Matthew Hackett, PhD,** is a Technical Lead and Manager for various medical simulation programs, including virtual patients, medical holography, serious games, mobile applications, TCCC, and volumetric at the U.S. Army CCDC Soldier Center. Hacket's research is focused on healthcare simulation and training, particularly military healthcare and combat casualty care.

**Jack Norfleet, PhD** is a Chief Engineer at the United States Army Combat Capabilities Development Command. He received his degree from the University of Central Florida. His interests are medical simulations and human tissue properties.

**Suvranu De** is the Google Endowed Dean for the FAMU-FSU College of Engineering and a professor of mechanical engineering. Dr De's research encompasses virtual reality, noninvasive neuroimaging, and artificial intelligence and their application to high-impact problems in healthcare.

# Airway Skill Assessment with Spatiotemporal Attention Mechanisms Using Human Gaze

**Jean-Paul Ainam, Rahul**
**CeMSIM, RPI**
**Troy, New York**
**ainamj@rpi.edu, rahul@rpi.edu**

**Lora Cavuoto**
**University at Buffalo**
**Buffalo, NY**
**loracavu@buffalo.edu**

**Matthew Hackett, Jack Norfleet**
**CCDC, Center STTC**
**Orlando, FL**
**matthew.g.hackett.civ@army.mil,**
**jack.e.norfleet.civ@army.mil**

**Suvranu De**
**Florida State University**
**Tallahassee, FL**
**sde@eng.famu.fsu.edu**

## INTRODUCTION

Airway management skills are essential in emergency medicine and are a required competency in various surgical education and certification programs. Traditional assessments of airway management skills heavily rely on subjective evaluations by trained professionals and are based on performing a fixed number of standardized maneuvers or procedures (e.g., Objective Structured Clinical Examinations (OSCE)) within a controlled environment. These conventional methods, though fundamental, often fail to gauge competency in real-world scenarios, particularly in emergencies where the ability to adjust to rapidly changing conditions is crucial. Recently, video-based assessment (VBA), a method of evaluation often conducted after the procedure, has received much attention (Maloney, 2021; McQueen et al., 2019; Pugh et al., 2021). VBA enhances patient safety during surgical procedures and enables surgeons to assess operations post-performance. However, watching an increasingly large number of video feeds and making accurate observations is challenging. Therefore, data-driven solutions that can process several video feeds simultaneously and make objective observations need to be developed.

In recent years, artificial intelligence (AI), particularly deep machine learning (ML) algorithms, have become more common and successful at handling video data (Funke et al., 2019; Yanik et al., 2022). These frameworks produce predictive capabilities using data and capture correlative relationships between input variables and the desired outcome via "training." Moreover, the choice of the ML model architecture, the quality and quantity of the training data, and the choices in executing the training loop primarily determine their performance. As a result, carefully designed ML systems can automatically and objectively detect the outcomes of a critical procedure in emergency medicine and save lives. In this paper, we explore the application of ML to multimodal data, *i.e.*, video and gaze, collected during an endotracheal intubation (ETI) procedure.

ETI is a procedure performed in emergency and critical care settings to establish and maintain a patient's airway. The ability to perform ETI swiftly and accurately is crucial, especially in high-stress environments, such as military operations. In these settings, the proficiency and rapid response of medical personnel can significantly influence the outcomes of life-threatening situations. In the military, soldiers are trained extensively in various medical procedures to provide immediate life-saving intervention on the battlefield or in other combat scenarios. ETI procedure involves inserting a flexible plastic tube through the mouth or nose into the trachea (windpipe) to establish an airway and ensure adequate ventilation, particularly in cases of respiratory failure or when a patient cannot breathe independently. This is crucial for maintaining oxygenation and preventing further complications. In combat, particularly, ETI may be performed by combat medics, corpsmen, or other trained personnel who are often the first responders in emergencies. They must quickly and accurately assess the patient's condition, determine the need for intubation, and execute the procedure under challenging and often austere conditions.

Given the critical nature of this procedure, ensuring its success is paramount. In this paper, we use an ML system to aid in this by ingesting various data types, such as the human gaze and video recordings of the procedure, to detect signs of successful or failed intubation. Through this system, military medical teams can improve the accuracy and efficiency of endotracheal intubation procedures, ultimately leading to better outcomes for patients in critical situations on the battlefield.

The ability to accurately recognize successful and unsuccessful ETI procedures in high-stress environments poses a challenging problem that requires an effective means of learning both spatial and temporal dynamics. Frame-based image cues have been widely used to recognize surgical activities; however, these models often depend on irrelevant features specific to the task environment rather than the surgical activities themselves (Sarikaya & Jannin, 2020). Using attention mechanisms with deep learning models has proven to significantly enhance performance in various tasks as they can focus on the more relevant features (Vaswani et al., 2017). These models learn an attention distribution, which is then used to adjust the weighting of features so that features that are more relevant to the task are given higher importance. On one hand, gaze tracking has been used to determine the feasibility of studying physician behavior during simulated medical emergencies (Szulewski & Howes, 2014). It has been shown that this technique is practical and useful and can provide new insights into clinician behavior and performance, which may lead to improved patient safety practices (Henneman et al., 2017). Consequently, the gaze data carry important information about the visual attention of humans and, thus, provide a natural means of an attention mechanism. They can be used to extract more relevant features that reflect these spatial and temporal dynamics in alignment with the human visual system. Deep learning models that utilize this information have proved effective in recognizing human activities, particularly in egocentric videos (Li et al., 2018).

This paper proposes using human gaze with a spatiotemporal attention mechanism for airway skill assessment in videos. The motivation is that during object manipulation, a significant portion of gaze fixation tends to be on the object parts pertinent to the task at hand. As a result, by directing attention to these specific regions, we can minimize extracting irrelevant features from cluttered backgrounds and objects unrelated to the ongoing task.

The proposed system is comprised of three steps. First, we model each gaze point in 3D space and design a visual mask with strong spatiotemporal correlations. The visual mask acts as patch selection and reduces redundant computation by discarding a proportion of irrelevant patches. To avoid errors introduced by gaze measurement, we corrected the visual mask by convolving an isotropic Gaussian over the measured gaze position. This ensures that absent or inaccurate gaze points are complemented, and different spatial locations remain observable even when the human gaze for the timestep is absent. Secondly, an autoencoder network is used to extract feature maps from videos, and an attention module is used to generate attention maps from the visual mask. In the last step, we combine the feature and attention maps and output a classification score. To our knowledge, this work is the first to propose using the human gaze to classify the outcomes of ETI procedures as successful or unsuccessful.

The contribution of the paper can be summarized as follows:
1. We incorporate a human gaze-guided spatiotemporal attention module in a model for clinical activity recognition. Our approach with human gaze outperforms the method without gaze.
2. We analyze the model's results using different performance metrics and demonstrate how the proposed method using human gaze contributes to better clinical activity recognition.

## METHODS

Modeling and automatically predicting clinical outcomes are fundamental for assessing psychomotor skills and providing timely feedback to trainees. In this paper, we propose a system that uses video and human gaze data to predict the outcome of a clinical procedure, specifically endotracheal intubation (ETI). The system consists of an autoencoder network that learns spatial dependencies directly from videos, producing a low-dimensional feature representation of the video. A set of pre-processing techniques converts the human gaze into a useful visual mask, which is then used by an attention module as a supervision signal to help the network selectively focus on salient parts of the video stream. Finally, a classifier outputs an outcome score. In the next sections, we describe each component.

### Visual Mask from Human Gaze

To guide the training of the spatiotemporal attention module, we create a visual mask from the human gaze as follows. We start by zero-initializing a grid plane $z \in \mathbb{R}^{H \times W}$, where $H$ and $W$ represent the spatial dimension of the input videos, and then model gaze locations as point coordinates in 3D space, *i.e.*, $\mathbb{R}^{T \times H \times W}$, where $T$ is the length of the video. At each time step, the gaze is fixated on a single location $(x_k, y_k)$ coordinate of the $H \times W$ space. Given a video frame, we enhanced its corresponding z plane by assigning a Kronecker delta function $\delta_{ij}^k$ as:

$$\delta_{ij}^k = \begin{cases} 1, & if\ (i,j) = (x_k, y_k) \\ 0, & otherwise \end{cases}. \tag{1}$$

This binary distinction is useful for filtering, masking, and selectively isolating or enhancing the influence of specific pixels in image and video processing. However, using a Kronecker delta function $\delta_{ij}^k$ to assign values based solely on matching gaze coordinates leads to sparsity in the resulting grid, with most pixels in the plane remaining zero. Consequently, data can be lost in the sea of zeros, making it hard for the model to directly use the human gaze to identify key features or differences across frames. To mitigate this issue, we conduct additional processing steps. We propagate the intensity of the initial pixel value (normalized to 1) to neighboring pixels based on their Euclidean distance from the central point. The propagation is governed by a decay function that reduces the influence as the distance increases until it reaches a predefined threshold, $\beta$. The propagation is defined as:

$$\delta_{ij}^k = ma\,x(\alpha^{d_k}, \beta) \quad if\ \alpha^{d_k} \geq \beta, \tag{2}$$

where $\delta_{ij}^k$ is the propagated value for the pixel at $(i,j)$. When $(i,j)$ matches the gaze coordinate $(x_k, y_k)$, this function is equivalent to Eq. (1), $d_k = \sqrt{(i-x_k)^2 + (j-y_k)^2}$ represent the Euclidean distance, $\alpha$ is the decay rate – a parameter less than 1 – that controls the rate of decay, $\beta$ is the lower threshold for propagation, ensuring values diminish with increasing distance.

We finally compute the visual mask for a given video by combining the masks from each gaze point as:

$$\delta_{ij} = \max(\delta_{ij}^1, \delta_{ij}^2, \dots, \delta_{ij}^n). \tag{3}$$

This ensures that the mask represents all gaze points effectively, highlighting areas with the highest propagated influence from any gaze point. However, this also create a mat of disjoint point center at the gaze location.

To fix this error and avoid errors introduced by gaze measurements and irrelevant glances, we apply an isotropic Gaussian convolution to the combined mask. This step smooths the mask, distributing the influence of individual gaze point more evenly and compensating for inaccuracies or absent data and is defined as:

$$G_{ij} = (G * \delta)_{ij}, \tag{4}$$

where $G$ is the Gaussian kernel and '$*$' denotes convolution. The Gaussian kernel G is defined as:

$$G(x, y) = \frac{1}{2\pi\sigma^2} e^{-\frac{x^2+y^2}{2\sigma^2}}, \tag{5}$$

where $\sigma$ is the standard deviation of the Gaussian distribution controlling the spread of the smoothing effect. The convolution process can then be written as:

$$G_{ij} = \sum_{u=-k}^{k} \sum_{v=-k}^{k} G(u,v)\delta_{i-u,j-v}. \tag{6}$$

Using Gaussian convolution to refine our visual mask with multiple gaze points mitigates measurement inaccuracies and ensures a broader visibility of gaze influence across spatial locations. This is particularly useful in ETI where gaze data must be robustly analyzed, ensuring that minor errors in gaze tracking do not adversely affect the overall data quality and usability.

Our pixel propagation method provides a balanced approach to enhancing gaze masks. It maintains a clear focus around the gaze point while sufficiently highlighting surrounding areas to avoid the pitfalls of sparsity. The method enhances the interpretability of gaze data in both low and high-resolution settings without significant computational overhead, making it suitable for real-time applications.

To unnormalize the grid into a visual mask (i.e., 8-bit images), we scaled the grid values back to the typical pixel value range using a simple linear transformation that adjusts the range of the data as

$$V_{ij} = \lfloor G_{ij} \times \kappa \rfloor, \tag{7}$$

where $V_{ij}$ represents the pixel value at the position $(i, j)$, $G_{ij}$ is the normalized pixel in the mask. $\kappa$ is the scaling factor (we use 255 in this paper), and $\lfloor \cdot \rfloor$ denotes the floor function.

Figure 1 shows the input image and the resulting visual mask with varying values of $\alpha$ in the first row. The second row shows the gaze heatmap, and the last row shows the masked video frames. The parameter $\beta$ is set to 0.25; its impact is marginal when transforming the grid plane to a visual mask (Eq. 7). The figure shows that the field of operation is captured by the visual mask. The mask locations accurately reflect the activity that is being carried out and focus on relevant parts.

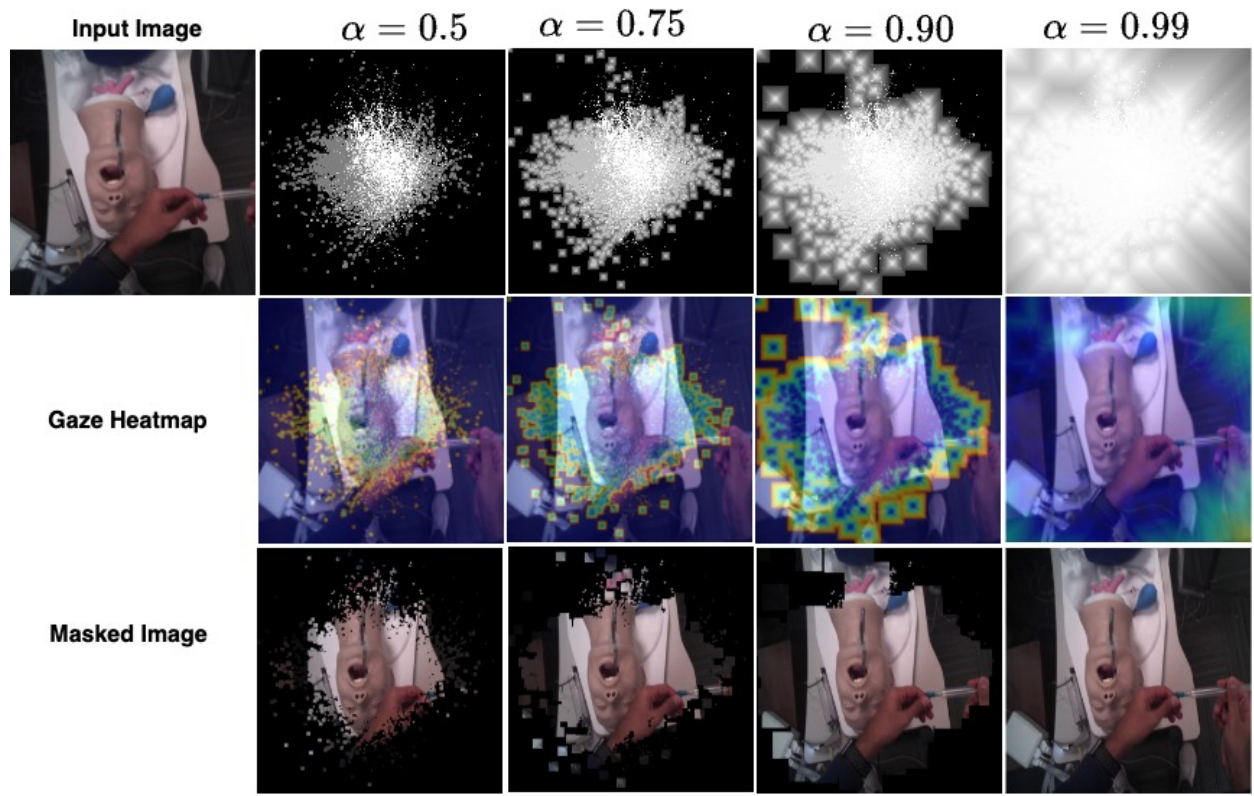

**Figure 1: Masking ETI procedure with the human gaze.**

Our masking approach design is strongly correlated with the spatiotemporal dynamic of the gaze locations and is deterministic. This allows for the generation of the visual mask prior to training the neural network classifier, thus reducing the overall training time. By adjusting the parameters $\alpha$ and $\beta$, we can modulate the granularity of the mask – either detailed or strict. A detailed mask preserves a larger area around each gaze point and reduces the risk of information loss, but it can potentially include irrelevant regions. Conversely, a strict mask confines attention to narrower regions, enhancing focus but risking omitting pertinent information. In this paper, we set $\alpha = 0.75$ and $\beta = 0.25$, striking a balance between overly sparse masks, which highlight only the exact gaze points, and overly dense masks, which obscure the distinction between the gaze points and their surrounding areas.

**Network Architecture**

The network architecture of the deep learning model is depicted in Figure 2 and is comprised of an autoencoder and a pre-trained self-supervised network.

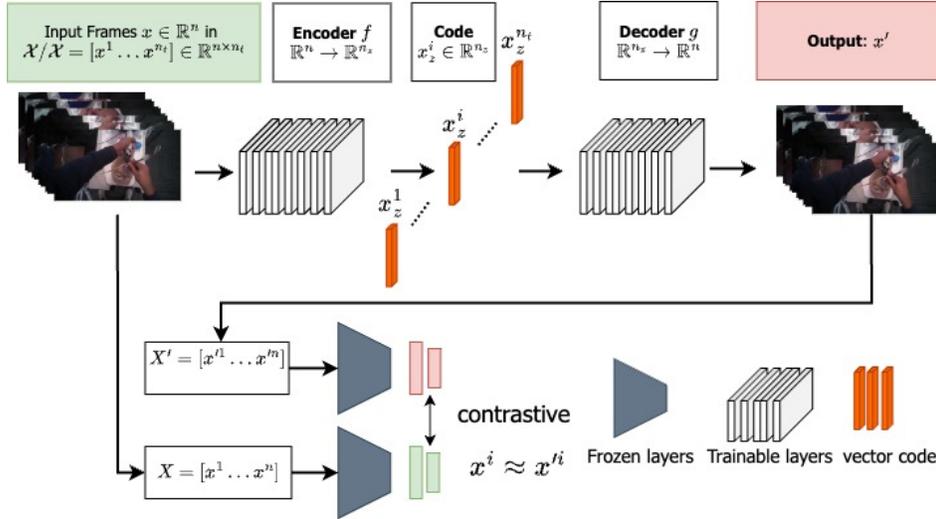

**Figure 2: The autoencoder framework. The network is trained on the original video and used to extract features from raw videos and visual masks.**

**Autoencoder**

An autoencoder (AE) is a neural network model that learns to encode input into a compressed latent space representation and then decode this representation back to the original input. We defined our autoencoder using two functions: an encoder $f$ that maps the input video $x$ to a latent representation $z = f(x)$ and a decoder $g$ that maps this latent representation back to a reconstructed input $x' = g(z)$. Figure 2 show the overall framework. This part of the network is trained to minimize the pixel loss (AKA reconstruction loss) and is defined as follows:

$$\mathcal{L}_{MSE} = \frac{1}{n}\sum_{i=1}^{n}\|x_i - x_i'\|_2^2\,,\qquad(8)$$

where $\|\cdot\|_2^2$ denotes the squared Euclidean (L2 norm). The term $x_i - x_i'$ represents the reconstruction error for the $i - th$ input sample. To encourage the autoencoder to capture key features of the input data, we enhanced the mean squared error (MSE) loss with a perceptual loss (Johnson et al., 2016). The perceptual loss uses a distance of visual features in lower layers as a measure instead of the original pixel-level comparison and is defined as:

$$\ell_{perc}^{\phi,i}(x,x') = \frac{1}{C_j H_j W_j}\left\|\phi_j(x) - \phi_j(x')\right\|_2^2\,,\qquad(9)$$

where $\phi$ is a pre-trained self-supervised network, whose parameters are frozen during the autoencoder training stage, $C_j, H_j, W_j$ are the channel, height, and width of the feature map of the $j$ convolutional layer, respectively. The final loss for the AE network is then defined as:

$$L_{AE} = L_{MSE} + L_p\,.\qquad(10)$$

## Attention Module

The attention module leverages the autoencoder features extracted from both the visual mask and the video sequence to guide the learning process of the classifier. The attention module and the classifier represent an end-to-end network based on the Squeeze-and-Excitation (SE) network (Hu et al., 2018). Our SE implementation consists of 1D convolutional blocks – a computational unit mapping an input $X \in \mathbb{R}^{C' \times L'}$ to feature maps $U \in \mathbb{R}^{C \times L}$. Let $V = [v_1, v_2, ..., v_C]$ be a learned set of filter kernels from a convolutional operator, where $v_C$ refers to the parameters of the $c - th$ filter. The output of the SE block is written as:

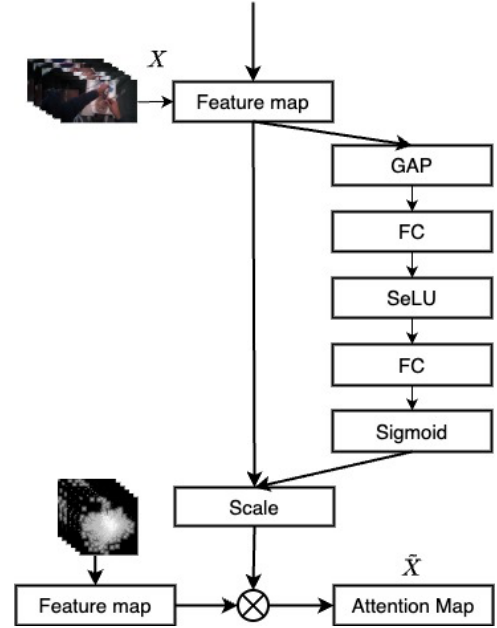

Figure 3: Attention module based on SE (Squeeze-and-Excitation) network.

$$u_c = v_c * X = \sum_{s=1}^{c'} v_c^s * x^s, \qquad (11)$$

where '$*$' denotes convolution, $v_c = [v_c^1, v_c^2, ..., v_c^{c'}]$, $X = [x^1, x^2, ..., x^{c'}]$, and $u_c \in \mathbb{R}^L$. $v_c^s$ is a 1D kernel representing a single channel of $v_c$ that acts on the corresponding channel of $X$. Here, bias terms are omitted to simplify the notation. The resulting feature maps $U$ is then element-wise multiplied by the feature from visual masks. This is regarded as fusion attention, where attention masks come from a different input modality (i.e., human gaze) and highlight features in the original video. This differs from self-attention, which uses features of a single input sequence to compute the attention scores. Figure 3 shows the attention module.

## EXPERIMENTAL DESIGN

### Datasets

The datasets for this study consist of video recordings and human gaze data of the ETI procedure, captured as part of a study approved by the University at Buffalo Institutional Review Board (IRB) and the US Army Human Research Protection Office (HRPO). Participants performed the ETI task on a standard airway manikin – the Life/form ® Airway Larry.

Video and gaze data was collected using Tobii Pro Glasses 2, a head-mounted camera system, recording at 30 frames per second. The video data includes 454 recordings from 18 subjects, comprising 325 successful and 129 unsuccessful trials. A head-mounted camera dataset presents unique challenges, such as significant head movement and unstable video quality, compared to datasets obtained from fixed cameras. Despite these challenges, the portability of head-mounted cameras makes them suitable for use in simulation centers and for training scenarios, such as for combat medics, without the complexities of setups involving fixed cameras. Head-mounted cameras capture critical visual aspects directly in front of the user, including gaze behavior.

In addition to video data, we collected human gaze data using eye-tracking glasses like a pair of eyeglasses to track the participant's eye movements. The glasses superimpose a gaze indicator on the first-person video, revealing where the participant is looking in real-time. Despite potential measurement errors due to blinking, rapid eye movements, and uncertainty in recording gaze fixation points, our proposed technique robustly accounts for such errors. Gaze data is originally represented as a two-dimensional coordinate, $(x, y)$ indicating gaze location within the viewer's field of vision and incorporates a temporal dimension reflecting the timing or duration of each gaze point. This three-dimensional approach — spanning spatial coordinates and time — provides a dynamic and comprehensive analysis of where and how long participants focus during procedures. This study specifically focuses on the data collected from the head-mounted cameras and the eye-tracking device.

For this study, participants were recruited through flyers distributed across the University at Buffalo campus. Each participant, who was required to be at least 18 years old and enrolled in a health-related major, received a $50 gift card as compensation and provided informed consent before participation. Additionally, they were given a one-to-one introduction to the study requirements.

**Implementation details**

The autoencoder network was trained for 100 epochs with a batch size of 128, using Adam optimizer (Kingma & Ba, 2015) and a learning rate of 0.001. We applied a dropout rate to 0.5. SimCLR was utilized as the pre-trained self-supervision network and fine-tuned on our dataset for 20 epochs (Chen et al., 2020). The classifier network, along with the attention module, was trained for 50 epochs using Adam Optimizer with a unit batch size and a learning rate of 0.002. The baseline model (M1) represents our system without the human gaze data, while the proposed model is referred to as M2 in the next sections. The framework was implemented using the PyTorch library (Paszke et al., 2019).

**Evaluation metrics**

We used several metrics to evaluate the efficacy of the models. Specifically, accuracy was used to assess the overall correctness, while the Matthews Correlation Coefficient (MCC) was used to evaluate the model's performance in the context of imbalanced datasets, offering a perspective that accounts for both positive and negative classes. F1-score, which measures the balance between precision and recall, was used to provide a single metric that highlights the model's precision in identifying relevant instances. Additionally, we calculated sensitivity and specificity to measure the proportion of actual positive and actual negative samples, respectively. Beyond these traditional metrics, we also investigated the trustworthiness of the two models.

**Trustworthiness**

Hryniowski et al., 2020; Wong et al., 2020 introduced a suite of metrics designed to evaluate the trustworthiness of deep neural networks. These metrics comprise question-answer trust, trust density, trust spectrum, and NetTrustScore (NTS), all of which rely on the SoftMax probability that reflects a model's confidence level $C(y|x)$ in its prediction $y$ given an input sample $x$. A model M is considered trustworthy if its true predictions are supported by stronger SoftMax probabilities. For any specific class $z$, the question-answer trust $Q_z(x, y)$, is defined as:

$$Q_z(x,y) = \begin{cases} C(y|x)^\alpha, & if\ x \in R_{y=z|M} \\ \left(1 - C(y|x)\right)^\beta, & if\ x \in R_{y \neq z|M} \end{cases}.$$ (12)

Here, $R_{y=z}$ and $R_{y\neq z}$ represents the spaces where predictions $y$ match and do not match ground truths $z$, respectively. The coefficients α and β serve as reward and penalty factors for true and incorrect predictions, and in this study, both are set to 1.

The trust density is represented as a probability density function over $Q_z(x, y)$ for each $z$, while the trust spectrum $T_M(z)$, aggregates the question-answer trust across all samples. This spectrum provides a range of trust values, offering insights into areas where the network is most and least trustworthy. It is defined as:

$$T_M(z) = \frac{1}{N} \int Q_z(x) dx \ ,$$ (13)

where $N$ denotes the sample size. The NetTrustScore (NTS) summarizes the overall trustworthiness of the model. It integrates the trust spectrum across all classes, providing a single value that represents the overall trust level of the model.

$$NTS = \int P(z) T_M(z) dz \ ,$$ (14)

where $P(z)$ is the output probability for ground truth $z$.

**RESULTS ANALYSIS**

**Table 1: Performance analysis**

|  | **Accuracy** | **MCC** | **F1-Score** | **Specificity** | **Sensitivity** |
|---|---|---|---|---|---|
| **M1: Without Gaze (w/o Gaze)** | 89.8 | 70.3 | 76.7 | 94.5 | 73.7 |
| **M2: With Gaze (w/ Gaze)** | 93.2 | 85.4 | 90.2 | 92.0 | 95.8 |

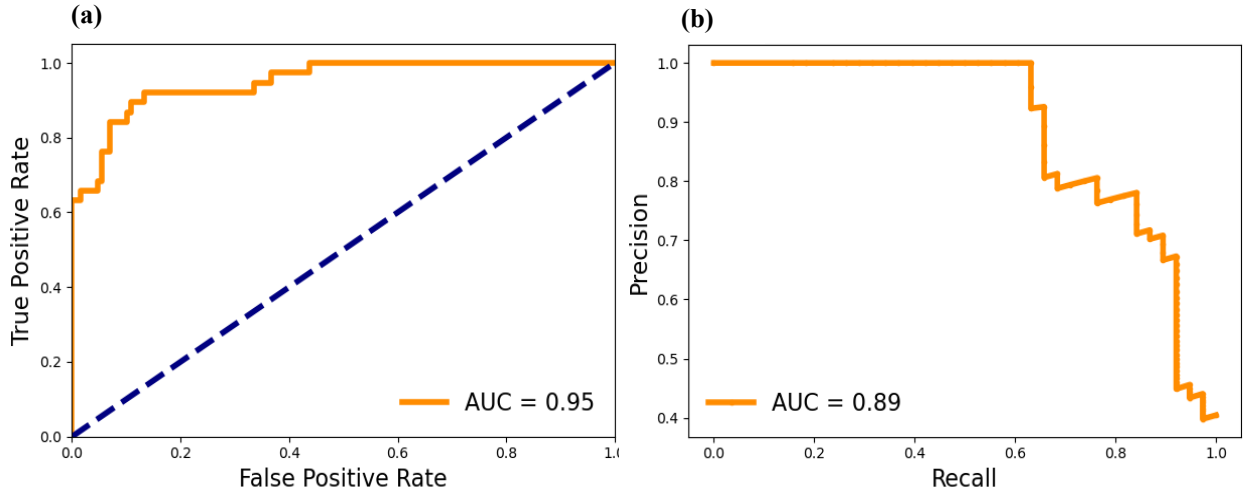

**Figure 4: ROC (a) and PR (b) Curves for w/o Gaze method.**

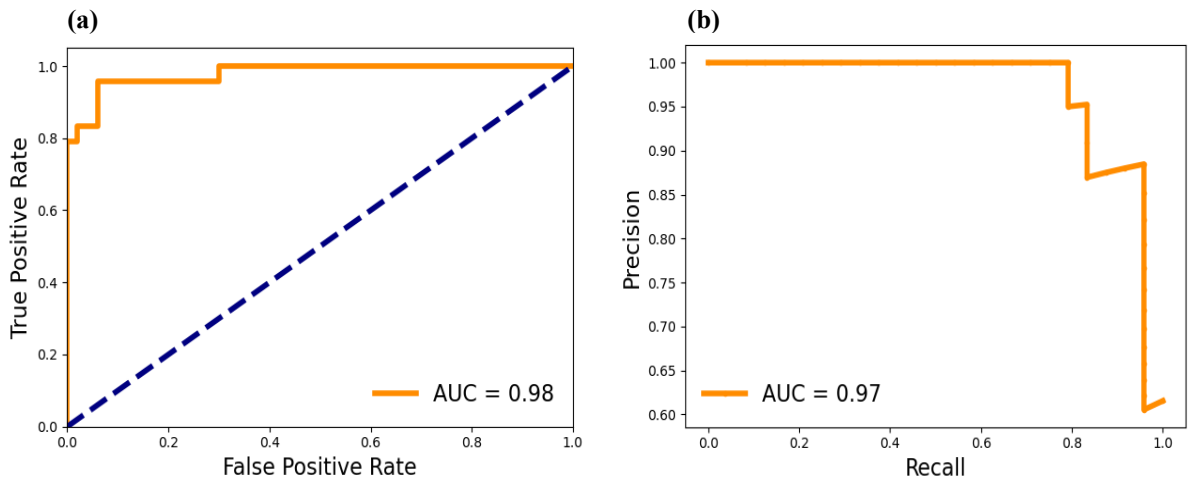

**Figure 5: ROC (a) and PR (b) Curves for w/ Gaze method.**

We compared the performance of the two models, M1 and M2, where M1 served as the baseline and M2 incorporated human gaze data. The integration of this additional data in M2 appears to have enhanced model performance across various metrics, as illustrated in Table 1.

The accuracy of M2 was observed to be 93.2%, an improvement over the 89.8% accuracy of M1. This suggests that using human gaze data contributes to the overall correctness of predictions made by M2. The MCC further supports this, with M2 achieving an MCC of 85.4 compared to 70.3 for M1, indicating a superior quality of prediction and a stronger correlation between the observed and predicted classifications in M2. Additionally, the F1-score, which assesses the balance between precision and recall, shows that M2's score of 90.2 substantially exceeds the 76.7 scores observed in M1, indicating not only better accuracy but also greater completeness in the prediction of positive class labels by M2. Specificity and sensitivity also reveal that while M2 has a slightly lower specificity of 92.0 compared to 94.5 in M1, it considerably excels in sensitivity, achieving 95.8 compared to 73.7 in M1. This higher sensitivity rate in M2 indicates a superior ability to correctly identify true positives.

We also report the areas under the curve (AUC) for both the Receiver Operating Characteristic (ROC AUC (a)) and Precision-Recall (PR AUC (b)) for M1 and M2 in Figure 4 and Figure 5, respectively. ROC AUC and PR AUC metrics in M2 are 98 and 97, respectively, compared to 95 and 89 in M1. These results confirm that M2 is more effective in discriminating between classes and maintains high precision across varying levels of recall.

Consequently, our strategic integration of the human gaze in M2, not only improves its performances relative to the baseline model M1 but does so across all examined metrics. This enhancement is particularly notable in the areas of sensitivity, MCC, F1-score, and AUC measurements, underscoring the significant impact of incorporating human gaze data into predictive modeling.

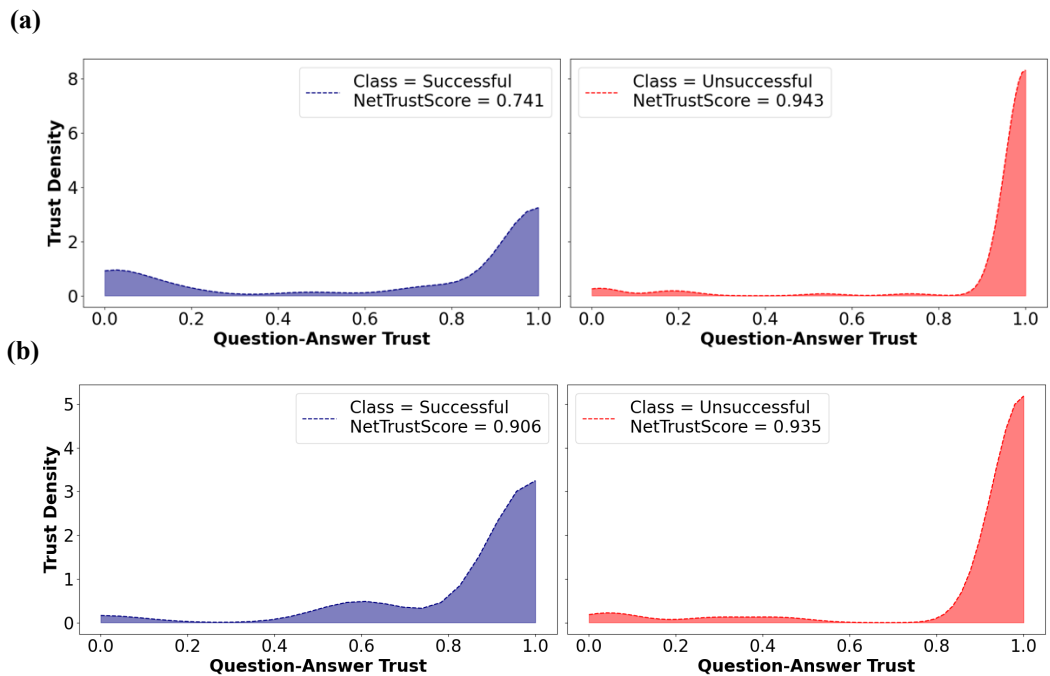

**Figure 6: Trustworthiness spectrum and NTS for M1 (a) without gaze and (b) with gaze.**

**Trustworthiness of the model**

The proposed system assesses clinicians' and medics' performance, removing potential biases and inconsistencies from the evaluation process. For the model to gain acceptance within the clinical community, it must be trustworthy, meaning its predictions must be reliable. This must be represented by high NTS scores (above 0.8) and a distribution skewed towards higher probabilities. Figure 6 illustrates the trust spectrum, the density of softmax probabilities for each class, and the area under curve values known as NetTrustScore (NTS).

The figure shows that the model with human gaze (M2) has a higher NTS for the successful class with a score of 0.906 compared to 0.741 for the baseline model (a). This indicates that M2 is more trustworthy when predicting successful classes, suggesting a stronger confidence level in its predictions for this class. Conversely, both models perform comparably in the unsuccessful class, with model M1 scoring slightly higher at 0.943 compared to 0.935 for model M2. The difference here is minimal, suggesting similar levels of trustworthiness for unsuccessful class predictions. Overall, using the human gaze improves the network's trustworthiness.

## LIMITATIONS AND FUTURE WORK

This study is part of a larger research initiative aimed at predicting the outcomes of clinical procedures using multimodal data. While the findings related to human gaze and video are significant, several limitations must be acknowledged.

Firstly, the modeling pipeline is not entirely automated and involves multiple steps. This setup might hinder seamless integration with existing techniques, highlighting the need for a fully automated, end-to-end learning model.

Secondly, gaze data was collected in a controlled environment and contributed to evaluating the procedure outcome. However, to validate the importance of gaze data, it is essential to collect data from high-stress situations, such as during combat.

Thirdly, previous studies have demonstrated that visual pattern is indeed associated with expertise (Gegenfurtner et al., 2017; Szulewski et al., 2018), but this study did not conclusively show whether gaze data alone can accurately assess a clinical procedure.

In future research, we will investigate whether visual difference in gaze data is a quantifiable measure of skill differences between novices and experts in high-stress environments. Integrating these objective measures for evaluating learner proficiency will enhance the validity of subjective and objective approaches to assessing trainee skills, and ultimately improve the granularity of resuscitation assessment while decreasing risks to patients.

## CONCLUSION

Automatic assessment of clinical skills is vital, particularly in military contexts where rapid and effective response to medical emergencies can save lives. In this paper, we proposed introducing an attention mechanism based on the human gaze to address the limitations of traditional and subjective evaluation methods of video-based assessments of airway management skills. The proposed method enhances medical training, particularly for emergency procedures like endotracheal intubation (ETI), and proves superior accuracy and efficiency compared to the counterpart model without human gaze. The framework's potential is to improve soldiers' ability to detect successful and unsuccessful ETIs under pressure in combat situations. Consequently, the method not only boosts competence evaluation but also enhances overall mission readiness, promising a profound impact on survival rates and operational effectiveness in the field. Future efforts will extend this innovative approach across various medical training applications.

## ACKNOWLEDGEMENTS

We gratefully acknowledge the support of this work through the U.S. Army Futures Command, Combat Capabilities Development Command Soldier Center STTC cooperative research agreement #W912CG-21-2-0001.